\documentclass[10pt, a4paper]{article}

\usepackage[utf8]{inputenc}
\usepackage[T1]{fontenc}
\usepackage[english]{babel}
\usepackage{newtxtext}
\usepackage{geometry}
\usepackage{authblk}
\usepackage{graphicx}
\usepackage[colorlinks=true, allcolors=blue, breaklinks=true]{hyperref}
\usepackage{parskip}
\usepackage{xurl}
\usepackage[numbers,sort&compress]{natbib}
\usepackage{lscape}
\usepackage{longtable}
\usepackage{booktabs}
\usepackage{float}

\title{\textbf{Challenges for Generative AI in Legal Reasoning}}
\author[1]{Eljas Linna}
\author[2]{Tuula Linna}
\affil[1]{Tampere University, Finland (\texttt{eljas.linna@tuni.fi})}
\affil[2]{University of Helsinki, Finland (\texttt{tuula.linna@helsinki.fi})}
\date{}

\begin{document}

\maketitle
\begin{abstract}
\noindent Large Language Models (LLMs) are being integrated into professional domains, yet their limitations in such high-stakes fields as law remain poorly understood. In response, this paper introduces examples of critical challenges to the functioning of generative and other forms of artificial intelligence (AI) as reliable reasoning tools in judicial decision-making. The study deconstructs core requirements and challenges for AI, including the ability to select the correct legal framework across jurisdictions, generate sound arguments based on the doctrine of the sources of law, distinguish \textit{ratio decidendi} and \textit{obiter dicta} in case law, resolve ambiguity arising from general clauses like "reasonableness", manage conflicting legal provisions, and apply the burden of proof correctly. The paper then maps various AI enhancement mechanisms, such as retrieval-augmented generation (RAG), multi-agent systems and neuro-symbolic AI, to these challenges, assessing their potential to bridge the gap between the probabilistic nature of LLMs and the rigorous, choice-driven demands of legal interpretation. Furthermore, the paper sketches a path towards an evaluation framework, proposing that legal requirements be organized into normative, doctrinal, evidential, and technical categories, and subsequently operationalized into domain-specific, testable design obligations. The findings indicate that these techniques can address specific narrow challenges, but they fail to solve the more significant ones that remain, particularly in tasks requiring discretion and transparent, justifiable reasoning. Therefore, we advocate for a staged adoption, first capturing efficiency in simple cases with technology already available today and thereafter sustaining long-term investment in new methods that handle hierarchy, temporality, and other requirements of legally sound reasoning, thus enabling expansion to complex adjudication in the future.
\end{abstract}



\section{Introduction}
Large Language Models (LLMs), machine learning systems that produce remarkably human-like responses, are rapidly being integrated into a wide array of professional domains \cite{Ling2024}. When assessed externally, these anthropomorphic artificial intelligence (AI) tools can appear to discuss, explain, and reason like human experts. This capability is fascinating, useful, and entertaining, but also fraught with risk when used carelessly. Shortly after the launch of the original ChatGPT tool in November 2022, embarrassing examples emerged of advocates who had used artificially manufactured legal citations \cite{Williamson2025}, most famously in the case of \textit{Mata v. Avianca, Inc.} \cite{Mata2023}. Today, we understand well that LLMs have severe limitations in their ability to reason and produce truthful responses. What remains less well understood, however, is where the boundaries of these limitations lie across domains.

This paper is intended primarily for researchers working at the intersection of AI and law, as it presents examples of challenging scenarios in judicial decision-making on a general, yet identifiable, level. Here, the standards for reasoning are not abstract ideals or judicial theories but concrete requirements for determining people's rights and obligations. Concerning AI in the legal context, it is useful to distinguish, at least, the following five aspects: (1) the regulation of AI, (2) AI-assisted research in legal theory, (3) AI's impact on access to justice, (4) AI-supported legislative drafting, and (5) AI-assisted or automated legal decision-making.

First, the \textit{regulation of AI} encompasses many important legal questions, such as copyright and privacy. Legal and ethical considerations have recently been raised about the methods of acquiring the training data for AI models, and about the application of results generated by these systems \cite{Pasetti2025}. For example, \textit{The New York Times} has already sued OpenAI and Microsoft for copyright infringement \cite{NYTvsOpenAI}, in response to which international bodies and national administrations are taking action: the European Union (EU) has passed Regulation (EU) 2024/1689 \cite{EUAIACT}, the first binding AI regulation in the bloc, while in the United States, action plans and executive orders have been issued to steer AI development \cite{WhiteHouseAI}. As such, this paper assumes such legal problems will be regulated in the coming years and thus omits them from the herein discussion.

The second aspect, \textit{legal theory and deontic logic}, represents an abstract normative science that studies the structure of normative systems and notions, such as "permissible", "obligatory", "optional", "must" and their counterparts. Symbolic AI has also shown some promise when used in the context of such theoretical patterns \cite{Kant2025}, and novel approaches have been presented to using LLMs in the translation of legal texts into formal deontic logic representations \cite{Horner2025}.

Third, AI has a significant impact on \textit{access to justice}. LLMs themselves are unlikely to create a competitive advantage among law firms, as all companies have access to the same leading commercial LLMs and open-source models. However, large law firms may gain an edge with proprietary data, especially in arbitration proceedings, where the datasets are not public. A vast proprietary knowledge base allows practitioners to utilize such techniques as retrieval-augmented generation (RAG) to ground LLM responses using examples and reference cases, leading to higher-quality responses from LLMs and thus higher-quality outputs. Conversely, AI can dramatically diminish the cost of legal aid; it has been noted that in high-volume litigation, a complaint response system reduced associate time from 16 hours to 3--4 minutes, and productivity gains of greater than 100-fold have been observed \cite{Couture2025}.

One emergent area of AI is in \textit{legislative drafting}, the fourth aspect. Because AI can streamline and optimize law-drafting processes, a cyclical interaction is enabled between LLMs and human legislative drafters, deepening the practitioners' understanding of the legal material and enhancing transparency in the law-drafting process. The EU has already started assessing the incorporation of AI into legislation editing software \cite{Fitsilis2024}.

This paper focuses exclusively on the challenges of applying AI in relation to the fifth aspect, \textit{AI-assisted or automated legal decision-making}, while contract interpretation is left aside, as there are special requirements compared to the interpretation of statutes and case law \cite{Arbel2024}. Our aim is not to test how reliably different LLM-based systems perform a given judicial task but rather to discuss the challenges that an AI system must overcome to succeed in the nuanced work of legal argumentation. The core question is what pitfalls AI must avoid and what new capabilities must still be developed for AI to serve as a reliable reasoning machine in the legal domain. 

To explore this question, this paper focuses on the process of legal reasoning from the standpoint of an impartial judge adjudicating according to the law. We use the Issue--Rule--Application--Conclusion (IRAC) model \cite{StockmeyerIRAC, Iijima1995} as our structuring device to avoid a mere list of challenges and to organize the discussion in a more coherent way. Thus, the IRAC model is not used as a theoretical model of legal reasoning or to depict how judges behave in real-life adjudication. It is merely a presentation technique to group some of the main challenges judges face in adjudication under different headings and to discuss the core questions of this paper.

The paper is structured as follows. Section 2 establishes a foundational framework for legal reasoning, outlining some of the core principles that typically govern judicial decision-making. Section 3 introduces some of the most challenging points in judicial argumentation for AI. Section 4 then maps the various AI enhancement mechanisms such as RAG, multi-agent systems and neuro-symbolic AI to these challenges, assessing their potential to meet the identified challenges. Section 5 outlines the steps towards a new framework for judicial AI, proposing a methodology to classify requirements into normative, doctrinal, evidential, and technical categories and to operationalize them as testable design obligations. Finally, Section 6 discusses the current state of AI in law, suggesting its most effective role today and looking towards a future in which human expertise and AI might work in symbiosis to uphold the rule of law.

\section{The Landscape of Legal Reasoning: A Foundational Framework}
\label{sec:landscape}
Before examining the specific challenges facing AI in the legal domain, it is essential to establish a foundational understanding of the requirements of the rule of law and a fair trial. We emphasize that the requirements presented below and the challenges mentioned in the next section are largely common, regardless of how legal argumentation is theoretically or practically constructed (e.g., as a logical, inductive or deductive syllogism, or construed according to, for example, moral-based Dworkinian interpretivism). For example, unclear wording of the law is a universal legal challenge.

At its heart, the judicial process is bound by a set of imperatives designed to ensure fairness and uphold the rule of law, and this set includes the principles of \textit{impartiality}, requiring decisions be made without bias; \textit{legal certainty}, ensuring outcomes are predictable and consistent; and \textit{transparency}, demanding the reasoning behind a judgment be accessible and understandable. Crucially, every step in the legal decision-making process must be \textit{justifiable and controllable}, allowing the parties to a dispute via appellate courts and the public to scrutinize and validate the outcome. These principles are not mere aspirations; they are functional requirements of a legitimate legal system. An important example of practical requirements is \textit{procedural economy}, i.e., how to implement the rule of law in legal proceedings cost-effectively. Here, as noted above, AI has the potential to save time and to speed up processes, but it may also have the opposite effect due to manual double-checking or correction of errors.

The IRAC model breaks down legal reasoning into four distinct phases:
\begin{itemize}
  \item \textbf{Issue (I):} Determining the core legal questions raised by the facts of a case.
  \item \textbf{Rule (R):} Identifying the relevant legal rules and case law that govern the issue.
  \item \textbf{Application (A):} Analyzing how the identified rules apply to the specific facts of the case. This is the heart of legal interpretation, where arguments are weighed and conclusions are drawn.
  \item \textbf{Conclusion (C):} Stating the outcome of the analysis and the resolution of the legal issue.
\end{itemize}

As noted, this paper uses the IRAC model as a technical framework for the discussion. It is not a depiction of real-world decision-making or an ideal model for high-quality legal argumentation. Rather, real legal argumentation is far more complicated than the IRAC model depicts. For example, IRAC lacks a dedicated letter for identifying relevant evidence and the specific facts of the case. When the letter F (Facts) is missing, the role of the relevant facts is discussed in the section concerning the application of rules (A).

While AI can be a powerful tool in the initial "Issue" discovery phase, its greatest challenges and potential lie in the subsequent stages. Judges have a legal obligation to deliver a judgment; a judge cannot say "\textit{non liquet}" (it is not clear), even when facing such significant challenges such as a gap in the law. Therefore, the two demanding phases of legal reasoning are the selection of the correct \textit{Rule (R)} and the nuanced \textit{Application (A)} of said rule to the facts. Here, in the interpretation and application of law, the highest standards of judicial reasoning are required.

Finally, alongside the interpretation of legal norms, the equally critical task of evidence evaluation forms another pillar of the judicial process. Here, we refer to the missing letter F in the IRAC model. A significant part of any trial involves assessing evidence---witness testimonies, documents, and forensic data---to establish a factual record. This process presents a distinct set of challenges for AI. Evidence evaluation is difficult to improve with techniques like RAG because legal evidence pertains to the truthfulness of everyday facts; fine-tuning an LLM with legal corpora does not help it answer whether a certain person was in a certain place at a certain time. As such, LLMs struggle to evaluate the truthfulness of real-world facts \cite{Quelle2024}, a task often reliant on human experience and credibility assessments. Furthermore, legal systems can employ complex and often unwritten \textit{burden of proof} rules to resolve cases in which evidence is insufficient. Thus, an AI system that attempts to decide a case based on weak evidence without correctly applying these rules would inadvertently bypass a cornerstone of procedural justice.

\section{Prominent Examples of Complex Legal Scenarios}
\label{sec:requirements}

The selection of rules, their application to facts, and the evaluation of evidence are not simple procedural steps but a series of complex legal choices that require deep domain knowledge, nuanced interpretation, and the ability to navigate ambiguity. This section discusses five core judicial challenges that AI systems must overcome to function reliably in legal decision-making. The purpose of this section is, on a general level, to map the challenges and pitfalls that are difficult to recognize when developing AI solutions in the legal domain. In different legal systems, these challenges manifest in more nuanced and concrete ways. Nonetheless, we believe it is important to map the challenges to ensure attention is paid to potential legal pitfalls to avoid a miscarriage of justice when relying on the help of AI in the legal domain.

\subsection{Selecting the Correct Legal Framework}
\label{subsec:select_framework}
Before any legal question can be answered, the correct body of law must be identified. However, this initial selection process is fraught with complexities, particularly in cases with cross-border implications or intricate national legal systems.

\subsubsection{Navigating International and Cross-Border Jurisdictions}
\label{subsubsec:cross-border}
In an increasingly globalized world, legal disputes often transcend national borders. Similarly, AI must first be able to distinguish between the \textit{jurisdiction} (which court has the authority to hear a case) and the applicable law (which country's laws govern the dispute). This distinction is a foundational concept, as the former (jurisdiction) is a part of procedural law, while the latter (applicable law) belongs to private international law. Traditionally, these two inquiries are treated as separate and sequential stages: a court must first determine whether it has the authority to adjudicate the matter before it proceeds to the second stage of deciding which substantive law to apply. This separation is particularly pronounced in legal systems that employ rigid, non-discretionary rules for each jurisdiction, such as those within the EU, where a court, once its jurisdiction is established, generally has no discretionary power to decline it on the grounds that a foreign court might be more appropriate (\textit{forum non conveniens}) \cite{CornellFNC}. Correctly distinguishing the two sides requires the proper application of international instruments, such as the EU's \textit{Regulation (EU) 1215/2012 (Brussels I Recast)} on jurisdiction and \textit{Regulation (EC) 593/2008 (Rome I)} on applicable law.

In addition, when determining the jurisdiction of the court, AI must identify whether a case falls under a \textit{uniform procedure} under EU law or some other supranational regulation where the procedure is the same in all countries in question, as well as whether the procedure is optional. Notable examples are \textit{the European Order for Payment Procedure (EC) 1896/2006} and \textit{the European Small Claims Procedure (EU) 2015/2421}. These regulations are optional and have their own jurisdictional provisions, giving national laws the power to regulate which court is competent. For example, in Finland, both procedures fall under the exclusive jurisdiction of the Helsinki District Court. 

Thus, in an EU country, AI must first decide whether the claim is of a national or international nature. If it is international, the AI system must decide whether EU regulation is applicable or whether the competent court is situated in a third country. If EU law applies, AI must choose between the Brussels I Regulation and the above-mentioned special EU procedures and then choose the competent court, while also considering that, for example, some provisions of the Brussels I Regulation and the special EU procedures allow national laws to regulate the competent court within that country. If there is a third country in question, AI must be able to apply the often open-textured and discretionary rules concerning connecting factors.

Furthermore, AI must recognize the fundamental differences between the two main legal traditions of common law and continental systems. This paper does not compare these systems but instead focuses on the challenges that solutions depending on AI must resolve before automation can be possible. Within the common law framework, in addition to written law, AI must respect the binding effect of precedent under the doctrine of \textit{stare decisis}. Moreover, AI must recognize how strong and strict the \textit{stare decisis} doctrine is in that country. In this regard, among others, the following questions must be considered:
\begin{itemize}
  \item whether all Supreme Court rulings bind all lower courts
  \item whether decisions of other appeal courts bind district courts
  \item whether \textit{stare decisis} binds across court lines (general/administrative/family, etc.)
  \item whether the Supreme Court is itself bound by its own precedents
  \item what is the relationship between the highest state and federal courts?
\end{itemize}

It is also worth noting that common law systems recognize various doctrines of non-binding or persuasive precedents \cite{Rachlinski2024}. While there is no formal rule of binding precedent in continental systems, many times, a consistent line of judicial decisions (\textit{jurisprudence constante}) from superior courts carries highly persuasive weight and is a crucial source for interpretation \cite{Fon2004}. Thus, AI must also find relevant precedents in continental systems.

In some domains, like international arbitration, it may even need to reason from non-codified sources, such as established trade practices (\textit{lex mercatoria}), or based on fairness (\textit{ex aequo et bono}). In these cases, the challenges facing AI are especially prominent.

While knowledge retrieval systems can find the relevant regulations, AI systems may struggle to interpret the connecting factors that determine jurisdiction, such as a defendant's habitual residence or where the damage occurred, which normally allow for judicial discretion and context-sensitive application \cite{Guitton2025}. A multi-agent system could break the problem into sub-tasks (agent 1 finds jurisdiction, agent 2 finds applicable law, agent 3 reviews, etc.), but this still relies on the agents' ability to interpret the often ambiguous and complex web of logical dependencies correctly.

\subsubsection{Respecting National Legal Hierarchies and Rules}
\label{subsubsec:hierarchies}
Even within a single country's legal system, choosing the right provision is a multi-layered task. An AI system must thus be able to navigate several critical dimensions:

\begin{itemize}
  \item \textbf{Hierarchy:} It must respect the hierarchical system of legal norms, prioritizing international treaties and constitutional law over ordinary statutes and decrees. Legal systems are structured with a clear hierarchy of norms, where rules from a higher source prevail over those from a lower source. An AI system must identify, for example, the relationship between international treaties and constitutions, as well as the relationship between federal and state laws. It must also recognize subordinate legislation, such as decrees and regulations. In the EU context, this includes understanding the difference between the \textit{direct applicability} of EU Regulations and, conversely, the interpretative effect of EU Directives on national legislation.
  \item \textbf{Time:} AI must correctly handle the \textit{time dimension}. Because laws are frequently amended, the version of the law that was in force when the events in question occurred must be applied. This concept, known as the temporal scope of law, is fundamental to legal certainty and its general principle is that of non-retroactivity. This requires checking for \textit{transitional provisions} that may suspend or modify the application of new laws and correctly identifying whether a law applies retroactively. Non-retroactivity is especially important in criminal cases (\textit{nullum crimen sine lege}), meaning an act can only be an offense if it was defined as such by law at the time of the offense. Transitional provisions can be complex when they bring some provisions into force immediately after the law has been ratified and the rest with different lengths of transition periods.
  \item \textbf{Specificity:} AI must correctly apply the general maxim of \textit{lex specialis derogat legi generali}, whereby a special law (e.g., a consumer protection act) supersedes a more general one (e.g., a general trade act). This principle is not used to solve contradictions but is applied to ensure the rule tailored to the specific situation is the one that governs it. This is not a simple keyword match; it requires the AI system to first correctly identify the legal status of the parties involved (e.g., consumer, employee) to determine whether a special law is applicable. The choice is simple when there is a single decisive factor, such as whether the plaintiff or defendant is in a particular position (such as a "consumer") and the laws in question are completely exclusive of each other. However, the relationship between general law and special law can be considerably more complex. Legislators tend to "conserve regulatory text" by constructing the relationships between laws in such a way that some of the provisions of the general law are applied either as such or with certain exceptions or modifications in special legislation. The same special law may refer to several general laws or to another special law. The landscape of the relationships between laws is therefore quite layered and difficult for AI to perceive. Moreover, reference provisions with the proviso "where/as applicable" (\textit{mutatis mutandis}) introduce their own nuances. These kinds of reservations give the applicant more leeway than a direct reference without any reservations. It remains to be seen how AI can account for such nuances.
  \item \textbf{Procedural Choice:} Within a single statute, an AI system must be able to choose the correct internal proceeding. For example, in the EU, some national insolvency acts contain preventive restructuring frameworks and discharge of debt (see EU Directive 2019/1023). Some insolvency laws also cover liquidation, whether as a going concern or via piecemeal sale. The U.S. Bankruptcy Code (Title 11 of the United States Code) is an example of this kind of multi-purpose insolvency act. Moreover, in the United Kingdom and in some other common law countries, there are company law-based procedures called "schemes of arrangement", where an AI system must be able to select the appropriate procedure based on the goals of the laws or different purposes of the chapters of the same law, such as rescuing a viable business or implementing an efficient sale of the estate. This type of choice is not only judicial; it also requires an analysis of the condition of the business in question.
\end{itemize}

The above-mentioned examples present a systemic reasoning challenge. An AI system must simultaneously consider multiple constraints, a task that is difficult to implement with simple prompts. For instance, determining the applicable version of a law (time dimension) while also checking for a superseding special law (specificity) and its place in the legal hierarchy requires a complex dependency graph. A neuro-symbolic AI system, which combines LLMs with rule-based systems, could perhaps model hierarchies of statutes, but defining all the rules and their internal relations is a monumental undertaking \cite{Kant2025}.

\subsection{Generating Sound and Defensible Legal Arguments}
\label{subsec:generate_arguments}
\subsubsection{Interpretative maxims}
\label{subsubsec:interpretative_maxims}
Once the correct legal text is identified, the core task of interpretation begins. This is not a matter of deductive logic but of constructing a persuasive argument grounded in accepted sources of law, generally accepted legal principles and the rule of law. Thus, traditional symbolic AI is unable to perform interpretation tasks. That said, there are certain traditional and quite universal unwritten interpretation maxims \cite{Barnes2023}, one of which is \textit{exceptio est strictissimae applicationis} (exceptions must be narrowly interpreted). However, if the purpose of the exception is to protect, for example, a weaker party, this maxim can be questioned. Another maxim is \textit{lex specialis derogat legi generali}, the application of which, as we discuss a bit later, can be quite challenging. A third example of interpretative maxims is \textit{lex posterior derogat legi priori}, the idea being that newer rules indicate the normative system in force, even when an older piece of law has not been repealed. Nevertheless, this is true only when the scopes of the application of the laws are identical. Furthermore, the maxim \textit{maior continet in se minus} (that is, the greater includes the lesser) means that if someone has the right to do something, they also have the right to do a smaller part of it. Finally, \textit{expressio unius est exclusio alterius} means that if the legislator has listed the items, circumstances or facts exhaustively, an \textit{e contrario} conclusion must be drawn if the thing in question is not listed. Even so, sometimes it can be unclear whether the list is meant to be exhaustive. The aforementioned interpretative maxims can be helpful when integrated into LLM prompting and agent architecture, but these maxims alone are insufficient tools in legal adjudication. Moreover, one must remember that the mentioned maxims may have various nuances and strengths in different legal systems and traditions.

\subsubsection{Adhering to the Doctrine of the Sources of Law}
\label{subsubsec:adhere_sources}

Legal argumentation follows a generally accepted, though sometimes flexible hierarchy of sources \cite{Pino2021}. A reliable AI must thus be able to construct its reasoning by starting with the literal \textit{wording} of the provision and interpreting it in light of the \textit{legislator's purpose}. Usually, the judicial task is to ascertain and act upon the "true intention of the legislature" \cite{Bell2018}; however, this is a potential pitfall, as the AI system must correctly source this purpose from preparatory works (e.g., government bills) and, crucially, verify whether Parliament amended the final law. If so, the reasoning in a parliamentary committee report supersedes the original preamble.

AI must then integrate relevant \textit{case law} and non-binding sources, such as the \textit{legal literature}, according to their authoritative weight as legal sources in that jurisdiction. This entire process must be conducted while respecting overarching legal doctrines, such as the \textit{principle of legality} and the \textit{rule of lenity} (\textit{in dubio mitius}) in criminal law. This is a problem of synthesis and validation, as an AI system must not only retrieve preparatory works but also verify whether the final law was amended, a step requiring a comparison of multiple documents and an understanding of the implications of any changes. This task demands more than simple information retrieval; it requires a structured comparison and logical inferencing about the significance of textual differences between legislative drafts, a higher-order reasoning capability that current LLMs often lack, leading to the risk of overlooking critical changes or hallucinating their importance \cite{Wu2024}.

\subsubsection{Interpreting Case Law}
\label{subsubsec:interpreting}
Using precedents requires more than simply finding cases with similar keywords. AI must be able to perform a sophisticated analysis of case law by distinguishing the \textit{ratio decidendi} (the core, binding legal rule of a judgment) from \textit{obiter dicta} (incidental remarks that are not binding), a particularly difficult task, as the original court does not explicitly label them. As described above, AI must thus identify which precedents have a binding effect and which are non-binding case law relevant for adjudication. 

Furthermore, AI systems must be able to conduct a \textit{legally relevant comparison of the facts} between a precedent and the current case, but a precedent is binding only if the cases are legally similar in terms of the key facts or are at least sufficiently analogous on legally significant points, a form of analogical reasoning influenced by societal values that goes far beyond matching semantic similarities via RAG. Legal analogy is not a mechanical process of counting factual similarities but an evaluative judgment about whether the facts of two cases are comparable in \textit{legally relevant} ways. This determination of relevance requires the identification of an underlying normative principle that connects the two cases, a step that moves from the specific facts to an abstract rule. This is one of the most difficult tasks for AI because \textit{ratio decidendi} is an abstract legal concept, not an explicit piece of text. Current AI systems, which excel at identifying semantic patterns in text, struggle to make the independent, evaluative judgments needed to formulate or apply these normative principles \cite{Sunstein2001}.

\subsection{Resolving Ambiguity and Incompleteness}
\label{subsec:ambiguity}
Law is not always clear or comprehensive. Therefore, a key judicial function is to resolve ambiguity and fill gaps in the law, tasks that demand reasoning beyond a formal application of rules.

\subsubsection{Adjudicating with General Clauses}
\label{subsubsec:general_clauses}
Legislators often employ open-textured terms like "reasonable", "fair", or "unconscionable" to delegate discretion to the courts, allowing flexible, case-by-case adjudication. These provisions, known as \textit{general clauses} or \textit{open rules}, are the legal rules deliberately formulated in an imprecise manner. They empower judges to decide cases based on broad principles, such as good faith, fairness, or public policy, when the strict application of a specific rule would lead to an unjust result—or when a specific rule does not exist.

To apply such general clauses, AI cannot be a mere \textit{legal formalist} that rigidly follows rules while disregarding non-legal factors. An understanding of social norms, ethics, and common sense is required, knowledge that is not explicitly codified. These general clauses are especially challenging for LLMs and cannot be solved simply with better information retrieval, as there is no reference material for general clauses in every context, nor by agentic systems, because the definition of the open-textured terminology cannot be broken down into well-defined factors.

\subsubsection{Addressing Conflicting Provisions and Gaps in the Law}
\label{subsubsec:conflicting_provisions_gaps}

In cases of a \textit{lacuna in law} (a gap where no specific rule exists in written law or binding case law), AI must be able to reason via analogies from related provisions or to derive a solution from the general principles that underpin that area of law. Instead of analogy, in some cases, the court can dismiss the claim or petition if there is no provision available. This outcome may be justified especially when it comes to whether an obligation should be imposed on the defendant/respondent or whether the plaintiff has a certain right regarding the defendant/respondent. An obligation that is not grounded on a clear rule may hamper justice, legal certainty, and the protection of legitimate expectations and trust.

Adjudicating without the support of a rule requires a deep, systemic understanding of the entire legal framework. While an LLM can process vast amounts of text, it struggles to track the dependencies and interactions across many factors, such as laws and principles simultaneously \cite{Kim2025}, a critical point of failure for AI.

\subsection{Establishing the Facts and Performing Reliable Evidence Evaluation}
\label{subsec:evidence}
Distinct from interpreting legal norms is the foundational task of establishing the facts of the case, which are crucial, even during the initial identification of the issue (I) that needs to be resolved. Without preliminary knowledge of the relevant facts of the case, it is not possible to determine the applicable law either. The parties may disagree about what happened and how the matter should be assessed legally. Sometimes, the issues of evidence and law may be intertwined, where the issue of law determines what evidence is relevant, or vice versa, so that before the issue of law can be determined, the disputed facts must be clarified. During the court proceedings, evidence presented can lead to a situation where the applicable rules change. Thus, there is a continuous interaction between the facts and rules that can be difficult for AI to follow. 

Reliability and accuracy are of utmost importance in evidence evaluation, as countless other procedures in law depend on the facts of the case. Even though LLMs have shown promising early results in evaluating evidence, considerable challenges remain unresolved regarding reliability and explainability \cite{Xing2025}.

\subsubsection{Assessing Factual Truthfulness}
\label{subsubsec:truthfulness}
The ability to reason about law is meaningless if one cannot ground this reasoning in a correct understanding of the facts. LLMs face a fundamental challenge in evaluating the truthfulness of information, such as evidence \cite{Wang2024}, as their knowledge is derived from training data and provided context, not from lived experience or an innate sense of real-world plausibility. For example, whether a specific witness is credible or whether a sequence of events described in a document is physically possible often cannot be inferred from the textual case documents alone. Assessing witness credibility involves evaluating non-verbal cues, demeanor, consistency, and potential biases, factors that extend far beyond the analysis of a written transcript. While AI tools are being developed to assist lawyers in witness preparation by analyzing speech patterns, emotional tone, and even micro-expressions, these are assistive technologies incapable of making definitive judgments about deception or truthfulness \cite{Nordstrom2025}. Evaluating real-world plausibility therefore requires common-sense reasoning, an area in which LLMs continue to show significant limitations, as they lack the embodied experience and intuitive understanding of the physical and social world that humans possess \cite{Khamassi2024}. Evidence evaluation involves piecing together fragments of information with no authoritative source of truth available for the LLM's reference, increasing the risk of hallucinations \cite{Dahl2024}.

\subsubsection{Applying the Burden of Proof}
\label{subsubsec:burden_of_proof}
When the evidence presented is insufficient or contradictory, a case is not left undecided (\textit{non liquet}). Instead, it is adjudicated according to the—often unwritten—\textit{burden of proof} rules, which prescribe which party bears the negative consequences of an unproven fact. The burden of proof is a \textit{general concept} widely shared across most legal systems for resolving uncertainty when the facts of the case cannot be definitively established. The details, however, vary significantly by jurisdiction and by the type of law (civil, criminal, administrative law, and specialized fields like tax law). In some systems, for example, the "burden of persuasion" and the "burden of production" are separated. A general starting point is that each party has the burden of proving their claims, although there are exceptions concerning, for example, weak parties, such as consumers. In addition, courts can consider the difficulty of proving a negative fact.

In this context, an AI system must be able to navigate these critical rules in civil cases. Moreover, in criminal cases, it must understand the high standard of "beyond a reasonable doubt", or a similar concept indicating a high threshold, and acquit if the prosecution fails to meet it. An AI system that attempts to render a decision based on insufficient evidence, rather than correctly identifying the issue as "unproven" and applying burden of proof rules, would fundamentally undermine procedural justice. The legal concept of the burden of proof is, at its core, a sophisticated mechanism for managing epistemic uncertainty. To apply it correctly, a decision-maker must first be able to assess its own level of certainty and recognize when the evidence is insufficient to meet the required threshold. This points to a critical technical limitation of current models, as LLMs are notoriously poor at this form of self-assessment. Their confidence scores are often misaligned with their actual accuracy, a problem known as poor calibration, and models frequently exhibit overconfidence, expressing high certainty in answers that are factually incorrect 
\cite{Zhou2025}.

\subsection{Upholding Procedural Fairness}
\label{subsec:procedural_fairness}
Finally, the legitimacy of a judicial decision rests not only on the quality of its reasoning but also on the fairness of the process used to reach it. The concept of a fair trial is multifaceted and ambiguous; usually, it includes the right to be heard and the opportunity for the parties to otherwise present their case; judges' impartiality and integrity, combined with the rules of judicial disqualification; and the rules regarding timely notification. In a case of procedural or material misjudgment, the right to challenge the judgment is an important judicial safeguard, although it is not an absolute right in all systems.

A cornerstone of a fair trial is the principle that both parties must have an opportunity to be heard and to contest the evidence and arguments presented by the other party. It is not enough to calculate how much time each party has had the opportunity to present claims, defense or evidence. Rather, what is important is that there is a fair balance between the parties' opportunities to appear and provide input. Important safeguards are that the parties are represented by professional lawyers and that the judge's role in directing and managing the progress of a case is fair and sufficient.

The essential questions concerning procedural fairness are:
\begin{itemize}
    \item whether the use of AI to manage the process like a judge hampers the fair trial requirement
    \item whether AI can enable the discovery of any errors or deficiencies concerning procedural fairness
    \item how to prevent parties knowledgeable about AI from gaining an unfair advantage in AI-powered processes over those with less understanding.
\end{itemize}

A judge's case management proceeds according to an evolving determination of relevance and can be quite passive in cases subject to party disposition (i.e., dispositive matters where the parties may settle the case). There is no predetermined path to be followed throughout the civil or criminal procedure, but the judge must be able to react according to the events and developments during the proceedings. However, it is not excluded that AI can manage, for example, repetitive small claims proceedings using digital forms for submitting the claims and objections of the parties. In such proceedings, the parties could be allowed to decide for themselves whether they have had a sufficient opportunity to present their case. An important legal safeguard is the possibility to appeal to a human judge.

Another issue is whether AI can monitor procedural fairness, for example, when a party wants to challenge the adjudication and appeal to a higher court. Discovering these kinds of matters requires that the procedure be transparent.

Many of these challenges are exemplified by prompt engineering, a standard procedure in the use of generative AI systems and a critical skill where subtle changes in wording, context, or structure can lead to vastly different outcomes \cite{Adams2025}. With the exact same prompt, different models produce different outcomes depending on their training data \cite{Kolbeinsson2024}. In an adversarial system, these choices would be subject to intense scrutiny. The principle of procedural fairness dictates that if an AI's output is to be used as evidence or to support a judicial decision, the process that generated that output must be transparent and open to challenge. The specific prompt used to generate a legal analysis is no longer just a query; it becomes a piece of discoverable evidence, and the choice of a particular AI model is a methodological decision comparable to an expert witness selecting a specific scientific instrument for an analysis. This necessitates a new level of transparency, extending beyond the final output to the entire generative process. A reliable and fair AI-assisted process must therefore be transparent enough to allow for such adversarial probing, ensuring that its methodological choices can be justified and challenged.

\section{AI Mechanisms and Their Application to Legal Requirements}
\label{sec:mechanisms}
Having deconstructed the complex requirements of the judicial process in the previous section, we now survey the established and emerging mechanisms that seek to bridge the gap between AI's probabilistic output and the discussed demands of legal reasoning. This section serves as an overview of AI methods and maps them to the challenges identified earlier. While prompting techniques are a core mechanism in interacting with LLMs \cite{Sivakumar2024}, we do not discuss them here, as they are already well understood and represented in the legal literature. 

No single technique is a panacea; instead, a combination of approaches is required to achieve reliability, transparency, and fairness in AI-assisted adjudication. We outline below how each class of techniques can address specific legal reasoning needs, drawing on recent research developments and highlighting their drawbacks. The final mapping of techniques and challenges is summarized in Table~\ref{table:techreqmap}.

\subsection{Domain Adaptation and Factual Grounding}
Before an LLM can reason about the law, it must first have reliable access to legal knowledge. This requires techniques that either adapt the model's internal parameters to the nuances of the legal domain or ground its responses in external, authoritative sources, thereby enhancing factual accuracy and reducing the risk of hallucination.
\begin{itemize}
    \item \textbf{Fine-tuning:} This method continues the training of a pre-trained model to adapt it to the legal domain's specific vocabulary, style, and nuances. Fine-tuning with legal domain data has been demonstrated to produce "remarkable improvements in various metrics, including the newly proposed Statute Relevance Rate and Legal Claim Truthfulness" \cite{Hu2025}. It allows an LLM to obtain a deeper foundation in legal domain knowledge, and it is useful specifically for grasping the structure and typical language associated with \hyperref[subsubsec:general_clauses]{\nameref*{subsubsec:general_clauses} (\ref*{subsubsec:general_clauses})}. However, fine-tuning alone is not enough to resolve the requirements discussed due to the variable and open-textured nature of the legal context, which often lacks an authoritative single truth. It also requires a high-quality dataset and risks "catastrophic forgetting", which degrades the model's performance in other capabilities that were not the focus of the fine-tuning process \cite{Liu2024}.
    \item \textbf{Retrieval-augmented generation (RAG):} RAG is the foundational technology for grounding an LLM's responses in a factual context, and it is an active subject of research in the legal domain \cite{Pipitone2024}. It uses various methods to locate relevant information in an external knowledge base (e.g., statutes, case law) and provides the retrieved content to the LLM as context, making it a powerful tool in scenarios that require pinpointing relevant documents, cases, or other information from a large corpus. RAG is essential for such tasks as \hyperref[subsec:select_framework]{\nameref*{subsec:select_framework} (\ref*{subsec:select_framework})} and \hyperref[subsubsec:adhere_sources]{\nameref*{subsubsec:adhere_sources} (\ref*{subsubsec:adhere_sources})} which require collecting many relevant documents before a decision on, e.g., the right jurisdiction can be made. It also significantly reduces (though does not eliminate) hallucinations \cite{Magesh2025}. On the downside, its effectiveness is only as good as the quality of the retrieval system and the knowledge base content. In addition, the most common RAG methods rely on semantic similarity rather than legal similarity, which may cause them to overlook relevant documents.
    \item \textbf{Temporal \& authority-aware retrieval:} Recent advances tailor RAG for the legal domain by incorporating time-sensitive and authority-aware retrieval strategies. One approach is to use temporal knowledge graphs that index laws and cases by effective dates, ensuring the LLM references the correct version of a statute in force at the relevant time. This addresses the issue of law evolving over time by modeling each legal provision as a sequence of timestamped versions, allowing answers that are accurate to a specific point in time and thus avoiding temporal hallucinations \cite{Martim2025}. In parallel, authority-aware ranking of results can be implemented by encoding legal hierarchies and citation networks into the retrieval process. For example, court decisions can be ranked with heavier weight on precedential authority (a Supreme Court ruling outranks a trial court opinion), using such graph algorithms as PageRank on the network of case citations. Such adaptations help AI prioritize sources that a jurist would consider more authoritative, thereby aligning retrieval with the doctrine of stare decisis and the hierarchy of norms \cite{Ho2025}. Together, temporal- and authority-aware RAG strategies improve the relevance and reliability of the model's legal knowledge base, grounding its answers in the right law (both in time and in authority).
\end{itemize}

\subsection{Structured Reasoning and Deliberation}
A legally defensible argument requires not only access to but also the transparent and logical application of correct information, which can be supported by methods that extend beyond direct prompt planning and that instill this discipline by structuring the LLM's cognitive process. They guide the model to move from simple probabilistic text generation towards more systematic and multifaceted reasoning, thereby producing a more explicit and verifiable logical trail for legal scrutiny.

\begin{itemize}
    \item \textbf{Task decomposition and chained prompts:} This technique breaks a complex problem into a series of narrower sub-tasks which are then sequentially processed using a chain of carefully crafted prompts. This increases accuracy and reduces hallucinations by allowing the model to take more manageable steps, and it has been shown to improve performance in legal reasoning tasks \cite{Yuan2024}. This is effective for processes that have a clear, sequential logic; for example, it can be used in \hyperref[subsubsec:cross-border]{\nameref*{subsubsec:cross-border} (\ref*{subsubsec:cross-border})} by prompting the LLM first to determine jurisdiction and then to identify the applicable law in a separate step. This approach fails when a legal problem requires a holistic evaluation and when different factors must be weighed simultaneously, without the ability to break them down into narrower, sequential steps.
    \item \textbf{Tree of thoughts (ToT):} When confronting an open-textured standard (like determining what is "reasonable" in a negligence case), ToT methods prompt the model to explore multiple reasoning branches in parallel. AI generates various possible interpretations or arguments and then evaluates and amalgamates them, akin to a deliberating judge considering pros and cons. This yields a more comprehensive analysis by explicitly mapping out competing arguments and their merits, and such an approach is valuable for tasks like \hyperref[subsubsec:general_clauses]{\nameref*{subsubsec:general_clauses} (\ref*{subsubsec:general_clauses})} that require balancing interpretations rather than the retrieval of a concrete rule \cite{Yao2023}. Unfortunately, ToT still relies on the model's internal logic, thus only helping in scenarios where thorough reasoning (not external data) can lead to the answer. It is thus a complement to retrieval-based methods, not a replacement.
\end{itemize}

\subsection{Hybrid and Agentic Architectures}
While the previously mentioned techniques enhance a single model, some legal challenges require capabilities that a monolithic LLM cannot offer alone, requiring more complex systems.

\begin{itemize}
    \item \textbf{Neuro-symbolic AI:} In this hybrid approach, an LLM acts as an "observation engine" to extract facts and rules, which are then processed by a classic, rule-based symbolic AI. This adds consistency and controllability to decisions by combining the LLM's powerful language understanding with the deterministic logic of a rule-based system. For example, in addressing \hyperref[subsubsec:hierarchies]{\nameref*{subsubsec:hierarchies} (\ref*{subsubsec:hierarchies})}, the LLM could identify the jurisdiction and dates of two conflicting statutes, which are then handed off to a symbolic engine encoded with the maxim \textit{lex specialis derogat legi generali} (a specific law overrides a general law) to decide which statute prevails. Similarly, a symbolic module can enforce \textit{lex posterior} (a later law overrides an earlier one) when applicable, an approach that guarantees core legal principles are followed step-by-step, offering a transparent chain of reasoning. The downside is the knowledge engineering effort: the legal rules (and all their exceptions) must be painstakingly codified, which in complex domains is a monumental undertaking. Neuro-symbolic systems excel in well-defined subdomains (e.g., computing priority of laws or sentencing guidelines), but they struggle to scale to the full ambiguity of general legal reasoning without extensive human-crafted rule sets.
    \item \textbf{Multi-agent systems:} Instead of one model attempting everything, this approach deploys multiple specialized LLM-based agents that collaborate as a team. Each agent is assigned a particular sub-task or role in the reasoning process. One agent might focus on finding all potentially relevant jurisdictional rules, another on gathering substantive law for the case, and a third on cross-checking for conflicts or exceptions. Through an organizer or orchestrator, the agents pass information to each other, mimicking a panel of experts or an adversarial setup. This decomposition can handle highly complex problems by dividing them: for instance, in \hyperref[subsec:select_framework]{\nameref*{subsec:select_framework} (\ref*{subsec:select_framework})} for a transnational case, agents can separately determine jurisdiction, identify applicable law, and evaluate any public policy exceptions. Agent frameworks can also simulate an adversarial proceeding, e.g., a "plaintiff" agent and a "defendant" agent debating to stress-test arguments and thereby help with \hyperref[subsec:procedural_fairness]{\nameref*{subsec:procedural_fairness} (\ref*{subsec:procedural_fairness})}. Unfortunately, agentic systems are still in the early stages of research and, at the time of writing, are often unstable, as a single agent's hallucination can have a cascading effect across the whole system \cite{Guo2024}.
\end{itemize}

\subsection{Output Verification and Refinement}
Given the probabilistic nature of LLMs, their outputs cannot be accepted without scrutiny. The final and perhaps most critical step in ensuring reliability is the verification of the generated content. This requires methods that enable the system to review its own work automatically for factual consistency or to assess its own confidence in a response, providing a crucial safeguard against error and a necessary component for responsible deployment in legal practice.

\begin{itemize}
    \item \textbf{Structured self-evaluation:} This involves prompting the LLM to create a list of evaluation factors to determine whether its own response is consistent and truthful against the provided source material and then evaluating these factors individually. This method acts as an automated quality check for verifying that the model has correctly followed the hierarchies in, e.g., \hyperref[subsubsec:adhere_sources]{\nameref*{subsubsec:adhere_sources} (\ref*{subsubsec:adhere_sources})} by decomposing the task into smaller and more easily accomplished steps. Decomposition has been shown to improve performance \cite{Yu2025}, but it is not fully reliable, as untruthfulness remains a persistent problem with LLMs \cite{Ozis2025}. Self-evaluation is also unhelpful when the truth is not explicit in the source material but instead requires nuanced judgment.
    \item \textbf{Logit-based confidence scoring:} To apply the burden of proof properly, AI must first be able to recognize when it does not have enough information. By looking at the raw logit outputs of the generated sequence, one can mathematically determine the confidence level in its response, allowing the system to reject an answer if confidence is too low. This is a potential technical approach to the challenge of \hyperref[subsubsec:burden_of_proof]{\nameref*{subsubsec:burden_of_proof} (\ref*{subsubsec:burden_of_proof})}. By quantifying uncertainty, it allows the system to recognize when evidence is insufficient and to flag a matter as "unproven" rather than rendering a decision on weak grounds. However, this method quickly loses effectiveness as responses grow longer and may surface the biases of the model, e.g., if its training data had more case examples of certain ethnicities, this could reflect in the confidence scores.
\end{itemize}

\subsection{Procedural Fairness and Transparency}
\label{subsec:procedural_fairness_transparency}
Introducing AI into judicial decision-making raises vital questions of procedural fairness: the processes must remain transparent, accountable, and subject to being challenged to uphold the rights of litigants. System design must therefore include features that log the AI's decision process and that allow meaningful oversight. Key requirements include:
\begin{itemize}
    \item \textbf{Audit trails:} All inputs, outputs, and intermediate reasoning steps of the AI should be recorded: the prompt given to the model, the documents retrieved, and any self-evaluation outcomes form a \textit{discoverable record}. This log allows opposing parties and appellate courts to scrutinize how the AI reached its conclusion, much as one would examine a human judge's reasoning for errors. Ensuring such traceability is not just for debugging; it is a legal necessity if AI outputs are to be contestable evidence in court.
    \item \textbf{Disclosure:} The use of AI in a case must be transparent to the parties. Recent guidelines emphasize clear disclosure of an AI system's role in judicial decisions \cite{Rojas2025}. In practice, if an AI draft or recommendation influenced a judgment, the judge should inform the parties of this fact and possibly provide the AI's report. Full transparency allows a party to object to or seek clarification about the AI's involvement, helping prevent a "black box" from quietly determining outcomes.
    \item \textbf{Contestability and human oversight:} To safeguard due process, there must always be an avenue for a human decision-maker to review and, if necessary, override the AI's output. For example, if an AI suggests a particular ruling, the affected parties should be able to challenge the reasoning, prompting a human judge to re-evaluate the issue in full. In effect, AI can assist but not replace human judgment. Courts and regulators are calling for independent oversight bodies and audit mechanisms to evaluate AI decisions periodically for bias or errors \cite{Rojas2025}. This ensures accountability of the AI system: like any expert tool, it must answer to human legal standards. By logging its processes and remaining under human authority, an AI-assisted adjudication system can be designed to respect fundamental procedural fairness, maintaining transparency, the right to be heard, and the right to an impartial, explainable decision.
\end{itemize}

\subsection{Evaluation Metrics for Legal AI}
\label{subsec:evaluation}
To assess these AI systems rigorously, domain-specific evaluation metrics are being defined that reflect core competencies in legal reasoning. Traditional Natural Language Processing (NLP) metrics like Bilingual Evaluation Understudy (BLEU) measure the linguistic properties of a text in reference to another, rather than the meaning of the content itself. They are therefore insufficient for legal tasks, where it must be assessed whether the AI found the right law, followed legal reasoning norms, and provided justifiable answers. The following are examples of emerging metrics for legal AI, as reported in the recent literature:

\begin{itemize}
    \item \textbf{Statute recall (SR):} This measures the AI system's ability to retrieve or cite all the relevant statutory provisions for a given issue. In judicial practice, \textit{precise recall} of applicable statutes is essential \cite{Xu2025}. For example, if a problem invokes a contract law issue, did the AI identify every pertinent section of the contract code? Recent benchmarks, such as LegalHalBench \cite{Hu2025} and CLaw \cite{Xu2025}, include tasks to quantify statute recall, finding that even top LLMs often miss applicable law or cite repealed statutes in error. A high statute-recall score indicates the system can comprehensively gather the letter of the law, a fundamental baseline for reliable legal analysis.
    \item \textbf{Ratio decidendi identification accuracy (RDIA):} Tests whether the AI can pinpoint the binding legal rule (\textit{ratio decidendi}) in a judicial opinion, as opposed to \textit{obiter dicta}. In scenarios heavy on case law, identifying the ratio is crucial for applying precedents correctly. Prompt-based evaluations for this skill have recently been designed, such as in \cite{Buckley2025}, where a prompt strategy was developed to guide an LLM to extract \textit{rationes decidendi} from judgments. The accuracy metric here is whether the AI's extracted principle matches the one legal experts recognize as the case's core holding. Improving this metric is vital for common law applications of AI, ensuring the system bases its reasoning on the central precedent rather than peripheral commentary.
    \item \textbf{Provenance coverage (PC):} This evaluates the extent to which AI-generated statements are supported by cited sources, essentially measuring how well-grounded the AI's output is in the evidence or law provided. A legally useful AI system should not only avoid hallucinations but also tie each factual or legal claim to an authoritative provenance (statute, case, etc.). One could measure the percentage of sentences in the AI's answers that are backed by a citation to the record or knowledge base, where high provenance coverage means the AI's answers can be verified by the reader \cite{Xu2025}. This aligns with metrics like the \textit{Non-hallucinated statute rate (NHSR)}, which checks that every statute the AI system cites actually exists and is relevant \cite{Hu2025}. In essence, provenance metrics quantify the AI's transparency and factual reliability, which are critical to building trust in AI-generated legal analyses.
    \item \textbf{Burden of proof handling:} A more novel evaluation criterion is how the AI handles uncertainty and evidentiary gaps, reflecting the legal concept of burden of proof. For example, does the AI appropriately refuse to answer questions conclusively when information is insufficient (analogous to a case not meeting its burden)? One could construct scenarios in which some necessary facts are intentionally omitted and score the AI on whether it refrains from speculation. Alternatively, in a dispute with issues requiring different standards of proof (e.g., "preponderance of evidence" vs "beyond reasonable doubt"), AI could be tested on treating them differently. While formal metrics here are still being developed, early work has the AI compute internal confidence levels and flag answers with low confidence. A system that declines to answer or issues an "unproven" verdict when appropriate would score well on this metric, demonstrating respect for the limits of available proof in its reasoning.
\end{itemize}

While the metrics above provide a foundation for evaluating performance for AI systems in the legal domain, several cross-cutting measures support finer-grained assessment across tasks. 
\textbf{Evidence attribution quality (EAQ)} gauges how accurately each substantive claim is tied to a supporting citation or document, complementing Provenance Coverage. 
\textbf{Jurisdiction / applicable-law accuracy (JAA)} captures the system's precision in selecting the correct forum and governing law when multiple may apply, while
\textbf{Temporal validity accuracy (TVA)} measures whether the system cites the correct version of a statute or regulation in force at the relevant time, ensuring temporal consistency. 
\textbf{Lex-specialis selection accuracy (LSSA)} tests whether a model correctly prefers a specific statute over a general one when both could apply, reflecting doctrinal hierarchy, while \textbf{Abstention under low confidence (AULC)} quantifies how often the system properly refrains from issuing a conclusion when confidence falls below a predefined threshold, aligning with the legal burden-of-proof principle. 
\textbf{Expected calibration error (ECE)} measures the correspondence between predicted confidence and actual accuracy across outputs, indicating how well model uncertainty aligns with truth, and
\textbf{Procedural logging completeness (PLC)} evaluates whether required audit-trail elements---inputs, retrievals, parameters, and outputs---are consistently recorded in terms of whether it is transparent and challengeable. 
Together, these metrics cover factual attribution, cross-border reasoning, and procedural fairness dimensions critical to judicial AI evaluation.

\begin{table}[H]
\centering
\footnotesize
\caption{Mapping of judicial requirements and challenges (Sec. 3) to AI techniques (Sec. 4) with limitations and relevant evaluation metrics.}
\begin{tabular}{p{2.8cm} p{5.6cm} p{3.3cm} p{3.0cm}}
\hline
\textbf{Challenge} & \textbf{Technique} & \textbf{Limitations} & \textbf{Evaluation Metrics} \\
\hline
Selecting the correct legal framework &
Authority-aware, jurisdiction-tagged retrieval using legal hierarchies and connecting-factor classification. &
Incomplete metadata; cross-system bias. &
SR, JAA \\[22pt]

Conflict of laws &
Knowledge-graph RAG encoding EU instruments (Rome I, Brussels I) and fora relationships. &
Sparse multilingual coverage; contextual ambiguity. &
SR, PC \\[16pt]

Respect hierarchy, time, and specificity of rules &
Temporal- and rule-priority-aware RAG enforcing \textit{lex posterior} and \textit{lex specialis}. &
Requires versioned corpora; exceptions hard to model. &
TVA, LSSA \\[16pt]

Apply sources of law and interpretation maxims &
Neuro-symbolic modules formalize interpretative canons for deterministic reasoning checks. &
Rule-encoding effort; limited scalability. &
PC, RDIA \\[16pt]

Interpret case law and extract \textit{ratio decidendi} &
Case-entailment and ratio-extraction pipelines align reasoning with binding precedent. &
Ratios implicit; analogies fact-sensitive. &
PC, RDIA \\[16pt]

Resolve ambiguity (general clauses) &
Tree-of-Thought and multi-agent debate to expose competing readings of open-textured terms. &
Debate instability; possible convergence to error. &
EAQ, PC \\[16pt]

Resolve conflicts and gaps in law &
Argumentation frameworks balance principles and fill lacunae under encoded norms. &
Formalization complexity; pragmatic nuance missing. &
LSSA, PC \\[16pt]

Evaluate truthfulness of evidence &
Evidence-grounded RAG linking factual claims to verifiable sources. &
Residual hallucinations; incomplete evidence base. &
EAQ, PC \\[16pt]

Apply burden or standard of proof &
Confidence-calibrated generation abstains when uncertainty is high. &
Calibration drift under domain shift. &
AULC, ECE \\[16pt]

Procedural fairness and explainability &
Comprehensive audit logging, disclosure bundles, and human oversight. &
Transparency vs. confidentiality trade-off. &
PLC, EAQ \\

\hline
\end{tabular}
\label{table:techreqmap}
\end{table}

\subsection{Example Architecture}
Because the purpose of this paper is to highlight the challenges in the legal domain to inspire new solutions and future research, we avoid proposing specific solutions to maintain focus on the challenges. However, we present an illustrative and high-level example of how the AI techniques may be leveraged to highlight the connection between the challenges and mechanisms.

\subsubsection{Neuro-Symbolic AI for Small Claims}
\begin{figure}[H]
    \centering
    \caption{Simplified architecture diagram of the example neuro-symbolic AI system for small claims}
    \includegraphics[width=0.95\linewidth]{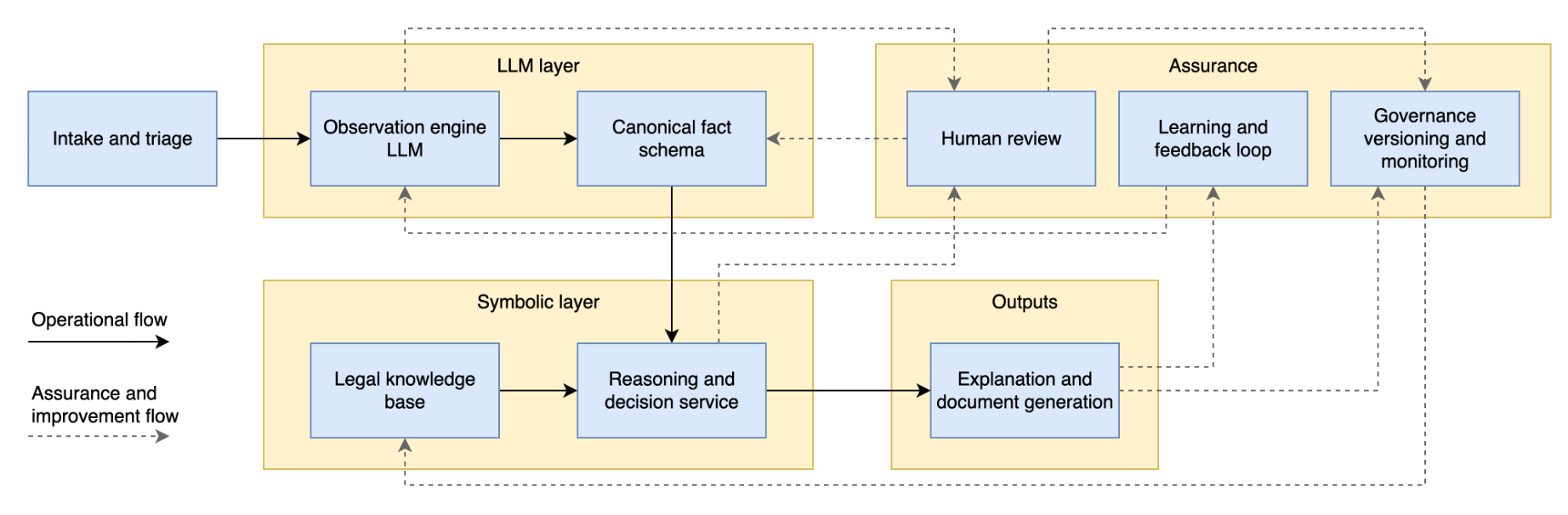}
    \label{fig:neuros-flow}
\end{figure}

A neuro-symbolic system is a feasible approach to resolving high-volume, low-complexity disputes, such as consumer product-defect small claims (e.g., under 10,000 EUR). The example system visualized in Figure \ref{fig:neuros-flow} assigns complementary roles: an \textbf{LLM layer} operates as an observation engine that reads unstructured inputs and proposes structured facts, while a \textbf{symbolic layer} performs the determinative legal reasoning by handling curated rules and decision tables. This division delivers both coverage over unstructured inputs and transparent, auditable decisions aligned with core legal maxims.

The pipeline begins with \textbf{intake and triage} (web form, document uploads, chat), which screens eligibility by subject matter, monetary threshold, and jurisdiction. The \textbf{observation engine} then extracts parties, amounts, dates, communications, and defect descriptions, and reconstructs a timeline into a \textbf{canonical fact schema} (e.g., in JSON). It attaches source citations and confidence scores, where contradictions or low confidence scores will route the case to review. These facts are submitted to a \textbf{reasoning and decision service} within the symbolic layer, which pulls jurisdiction-versioned statutes and decision models, such as limitation periods, notice and cure requirements, remedy ladders, or damages caps from the \textbf{legal knowledge base}. Conflicts among candidate norms are resolved by a submodule within the reasoning service that encodes \textit{lex specialis derogat legi generali} and \textit{lex posterior} using rule metadata (specificity, effective dates, jurisdiction) to select the controlling norm.

The reasoning service applies decision tables and forward-chaining rules to produce findings (e.g., timeliness, opportunity to repair), liability determinations, and permissible remedies, and it emits a decision trace, the ordered list of rules fired, and the facts they consumed. These are submitted to an \textbf{explanation and document generation layer}, which produces human-readable rationales and drafts, such as settlement letters and small-claims pleadings, parameterized by the symbolic outcomes, and populated with cited facts.

Quality is maintained through the \textbf{human-in-the-loop assurance layer} (a review console for fact corrections and re-runs), \textbf{governance} (versioned rules, audit logs, jurisdictional whitelists), and \textbf{monitoring} (extraction error taxonomy, disagreement and appeal rates, data drift). Thus, deployment proceeds in a bounded domain, expanding iteratively as knowledge engineering matures. In line with our dual-application model, this neuro-symbolic approach enables high-quality automation for simple cases with built-in safeguards, while preserving a clear escalation path to human expertise when ambiguity arises.

To measure system quality, we prioritize a compact set of metrics discussed in Section~\ref{subsec:evaluation}: SR, JAA, TVA, and LSSA, which together test whether the system selected the correct, in-force law and applied doctrinal precedence consistently. We complement these with PC to quantify the grounding of each conclusion in the record or corpus, and AULC to ensure appropriate non-decisions on under-specified cases. Tracking these metrics over versioned rule sets provides a concise, decision-relevant signal of legal correctness, transparency, and safety.

\section{Towards a Framework for Judicial AI Requirements}
\label{sec:towards_framework}

Due to the abstract and broad nature of the overview discussed in this paper, the paper stops short of proposing a full-fledged framework for AI-assisted adjudication. In this section, we outline steps towards defining such a framework, and we discuss how the heterogeneous challenges identified in previous sections can be classified, operationalized, and evaluated. The aim is to sketch a research program rather than to complete it within the confines of this paper.

As a first step, a framework for judicial AI should organize requirements into distinct categories to enable a more granular investigation. We propose the following categories: (1) \textit{normative and procedural values}, such as fairness, transparency, and legal certainty; (2) \textit{doctrinal and reasoning constraints}, such as the hierarchy and temporality of norms, sources-of-law doctrine, and interpretation methods; (3) \textit{fact-finding and evidential requirements}, including credibility assessment and burden of proof; and (4) \textit{system-level technical properties}, such as calibration, auditability, and logging. Table~\ref{tab:req_categories} offers an initial, high-level mapping of the requirements and challenges discussed in Section~\ref{sec:requirements} to these four categories. This categorization is explicitly presented as a preliminary suggestion: the requirements are interconnected, many items plausibly belong to more than one category, and their precise classification will vary across legal domains and jurisdictions. Future research should therefore refine and stress-test this mapping in narrower, more well-defined areas of law rather than seeking a universal taxonomy at once.

The second step is to \textit{scope these requirements to concrete legal domains}. Rather than aiming at abstract completeness, a pragmatic framework would be developed domain by domain, for example, beginning with small-claims consumer disputes or other high-volume, low-complexity matters. For each chosen domain, one would enumerate: (i) the core normative values that are engaged (e.g., protection of weaker parties, procedural economy); (ii) the relevant doctrinal structure (hierarchies of norms, special regimes for consumers, mandatory versus default rules, interpretation practices); (iii) typical evidential questions and associated burdens and standards of proof; and (iv) the system-level properties that any AI component operating in this domain must possess. This converts the general catalogue in Section~\ref{sec:requirements} into a domain-specific checklist rather than a universal theory of adjudication.

Third, for each requirement in this domain-specific checklist, the framework would specify an \textit{operational design obligation} for AI systems. For normative and procedural values, these obligations take the form of process guarantees: for example, transparency may be operationalized as a requirement that the system exposes, at a minimum, the key legal sources on which it relied, the main inferential steps taken, and any confidence or uncertainty assessments that influenced the outcome. Meanwhile, legal certainty may be operationalized as a requirement that the system systematically consult a versioned corpus of norms and annotate each citation with its temporal validity, ensuring the law in force at the relevant time was applied. For doctrinal constraints, operationalization often means encoding structural relationships: for instance, a requirement that the system enforce \textit{lex specialis derogat legi generali} in predefined clusters of norms, or that it distinguishes primary from persuasive authorities when generating a recommendation. Evidential requirements translate into obligations concerning how the system handles missing, conflicting, or weak evidence; for example, a requirement is that the system must be able to output that a fact remains "unproven" and to apply the burden-of-proof rules of the relevant legal order accordingly, rather than silently assuming a factual conclusion. System-level technical properties are specified as constraints on the architecture: logging of inputs, prompts, retrieval operations and outputs; human-in-the-loop oversight points; and the use of calibrated uncertainty estimates to trigger abstention, escalation, or explicit signaling of residual doubt.

The fourth step is to make these operational obligations \textit{testable}. Here, the techniques and metrics surveyed in Section~\ref{sec:mechanisms} become tools for turning legal desiderata into falsifiable engineering criteria. For doctrinal constraints, this involves designing benchmark tasks that probe, for example, the accuracy of jurisdiction and applicable-law selection (JAA), correct selection of the in-force version of a statute (TVA), consistent preference for special over general provisions where legally required (LSSA), and the ability to identify the \textit{ratio decidendi} of a precedent (RDIA). For transparency and controllability, such metrics as PC and EAQ measure the extent to which the system’s outputs are grounded in identifiable legal sources. For evidential and burden-of-proof requirements, calibration-oriented metrics and abstention measures (such as ECE and AULC) provide proxies for the system’s ability to recognize when the available information is insufficient to support a conclusion and to signal this insufficiency in a way that aligns with the legal structure of proof. For procedural fairness and auditability, one can define a procedural logging completeness score that checks whether required artifacts, such as prompts, model versions, retrievals, or timestamps, are consistently recorded and made available for oversight.

Crucially, these tests and metrics must be tied back to the underlying requirement type. The framework would explicitly state, for example, that a given small-claims decision aid is acceptable for deployment only if it meets minimum thresholds on a bundle of doctrinal metrics (e.g., SR and TVA), transparency metrics (e.g., PC), evidential metrics (e.g., abstention behavior in under-specified scenarios), and procedural metrics (e.g., logging completeness and availability of human review). In this way, RAG, multi-agent systems, and neuro-symbolic techniques are no longer merely cataloged under headings that echo the earlier requirements; they become components that can be combined to satisfy specific, classified obligations, and their adequacy can be evaluated with reference to explicit tests.

Finally, the framework must remain \textit{jurisdiction-aware and jurisprudentially modest}. It should record, for each doctrinal and evidential requirement, the extent to which it is stable across legal families and the points at which legal cultures, institutional arrangements, or interpretive theories diverge. Rather than asserting that "judicial reasoning is" or "must be" conducted in a single way, the framework would describe which families of adjudicative practice it is designed to approximate and which trade-offs this choice entails. Because requirements interact differently in different areas of law, subsequent work should focus on narrow domains one at a time, progressively refining both the categorization in Table~\ref{tab:req_categories} and the associated operational and evaluative criteria. This paper has provided the raw material for such an endeavor by mapping difficult points in judicial reasoning and aligning them with plausible AI mechanisms and evaluation tools. The steps outlined in this section indicate how this material could be reworked into a structured, domain-sensitive framework in subsequent research.

\begin{table}[H]
\centering
\footnotesize
\caption{Illustrative mapping of requirements and challenges (Section~\ref{sec:requirements}) to requirement categories. The classification is preliminary, and concrete applications are expected to refine it on a domain-by-domain basis.}
\begin{tabular}{p{3.1cm} p{10.4cm}}
\toprule
\textbf{Requirement category} & \textbf{Illustrative requirements and challenges from Section~\ref{sec:requirements}} \\
\midrule
Normative and procedural values &
Impartiality, legal certainty, transparency and justifiability of reasoning; fair trial guarantees (right to be heard, opportunity to contest evidence and arguments, appeal); protection of weaker parties (e.g., consumers); principles of legality and lenity; value-laden application of general clauses (reasonableness, fairness, good faith, public policy); legitimacy and restraint in gap-filling; equality of arms and avoidance of unfair advantage from AI use; contestability and openness of AI-assisted processes. \\[6pt]

Doctrinal and reasoning constraints &
Distinguishing jurisdiction from applicable law; applying EU instruments and special procedures; handling connecting factors and conflict-of-laws; respecting hierarchy of norms (constitutional, international, statutory, delegated legislation, case law); temporal scope of law, non-retroactivity, transitional provisions; \emph{lex specialis} and \emph{lex posterior} relations; selecting appropriate procedures within multi-purpose statutes; interpretative maxims; doctrine of sources (wording, legislative history, case law, literature, principles); precedent hierarchies; distinguishing \emph{ratio decidendi} from \emph{obiter dicta}; analogical reasoning between cases; methods for applying general clauses; gap and conflict resolution techniques (analogy, principles, dismissal where no legal basis exists); formal burden-of-proof schemas. \\[6pt]

Fact-finding and evidential requirements &
Establishing relevant facts as a precondition to identifying issues and applicable law; managing the interaction between evidential and legal questions; assessing plausibility and credibility of accounts; dealing with incomplete, inconsistent, or noisy evidence; applying burdens and standards of proof in civil and criminal matters (including reversals and difficulties of proving negatives); correctly treating issues as unproven when evidence is insufficient; ensuring that downstream legal conclusions track the applicable burden-of-proof rules rather than unsupported factual assumptions. \\[6pt]

System-level technical properties &
Temporal and authority-aware retrieval and knowledge representation; mechanisms to encode and enforce hierarchies, specificities, and precedence between norms (including neuro-symbolic modules); support for structured reasoning, self-evaluation, and abstention under low confidence; calibration and uncertainty estimation aligned with legal burdens of proof; provenance tracking and evidence attribution linking outputs to underlying sources; comprehensive logging of inputs, prompts, model versions, retrievals, and outputs; disclosure and audit interfaces; human-in-the-loop oversight and escalation paths; access control and interaction design that preserves fairness and contestability. \\
\bottomrule
\end{tabular}
\label{tab:req_categories}
\end{table}


\section{Conclusion}
The preceding sections have deconstructed the intricate process of judicial reasoning, pointing out that in complicated cases, the "Rule" and "Application" phases encompass tasks that are challenging for LLMs due to the wide discretion required on many levels. The analysis reveals gaps between the fundamental probabilistic nature of LLMs and the rigorous, choice-driven nature of legal adjudication. The crucial question is whether LLMs can make justified legal choices in accordance with choice-of-law rules, between several interpretation alternatives, or to fill a lacuna in law and, critically, express the reasoning chain, showing how arguments and counterarguments were weighed. Without this transparency, it is impossible to ascertain the due course of reasoning.

\subsection{The Evolving Symbiosis of Law and AI}
This paper has focused on the most challenging situations and aspects of law, a review that may seem overwhelming considering the broader potential of LLMs. However, many judicial cases are in practice rather simple. It is thus more productive to view AI not as a monolithic replacement for human legal interpreters, but as a powerful tool whose application must be carefully tailored to the complexity of the task. We suggest that the most effective current role for AI in law can therefore be understood through a dual-application model: as a high-volume assistant for simple cases and as a sophisticated "sparring partner" for experts in complex matters.

First, in simple high-volume legal cases where only a few points must be considered, LLMs can already work well with today's techniques. Matters such as small claims procedures or consumer disputes concerning product defects often feature repetitive fact patterns and are governed by an extensive body of case law. For example, small claims procedures with disputes between a consumer and a vendor concerning a product defect in goods are an ideal environment for systems such as neuro-symbolic AI, allowing high-quality automation with safeguards in place. If a party wants to challenge the response of the system, a human operator can easily validate the details extracted by the LLM and verify whether the symbolic logic is correctly applied.

Second, regarding most complex civil and criminal litigation, there arises a deep barrier and the fundamental mismatch between the distinctive character of legal reasoning and the probabilistic nature of the core generator, the LLM. Yet, even in these tricky cases, LLMs can assist legal decision-making by saving time in document management, identifying arguments, and enhancing efficiency in many ways. Here, the LLM is not an adjudicator but an invaluable collaborator for the human lawyer or judge, helping to stress-test arguments and enhance the coherence of a legal strategy.

\subsection{Synthesis and Future Directions}
Stated simply, the current generation of LLMs excels when legal tasks can be distilled into sophisticated information retrieval and pattern recognition. They serve best in cases that have abundant reference material, rely on clearly defined concepts as opposed to open-textured standards, and can be broken down into narrow, sequential sub-tasks. When these conditions are not met, a fundamental barrier remains: the gap between the probabilistic nature of language models and the principled, choice-driven nature of judicial reasoning.

This is not a poor outcome. We see that the successful automation of high-volume, procedurally simple cases has the potential to free up significant human resources for more complex legal challenges. A fairly substantial number of court cases falls into the category of simple and standard, or even identical, disputes with predictable outcomes \cite{Laptev2024}. In any case, the path towards broader application in complex litigation requires a focused effort to develop systems capable of justifiable reasoning. Future research needs to prioritize creating AI with auditable decision chains, reliable mechanisms for self-assessing uncertainty to respect legal principles, and hybrid architectures that can enforce the rigid hierarchies and non-negotiable rules of legal doctrine. Despite the challenges, we remain optimistic that the current trajectory of technological advancement will bring these capabilities within reach in the future.


\section*{Declarations}

\noindent\textbf{Funding} The authors declare that no funds, grants, or other support were received during the preparation of this manuscript.

\medskip
\noindent\textbf{Competing interests} The authors have no relevant financial or non-financial interests to disclose.

\medskip
\noindent\textbf{Ethics approval} Not applicable. This study is purely theoretical and analytical in nature and did not involve human participants, animal subjects, or personal data.

\medskip
\noindent\textbf{Consent to participate} Not applicable, as the study did not involve participants.

\medskip
\noindent\textbf{Consent for publication} Not applicable, as the study does not include personal data.

\medskip
\noindent\textbf{Availability of data and materials} Not applicable; no new datasets were generated or analyzed during the current study.

\medskip
\noindent\textbf{Authors' contributions} Both authors contributed equally to the research and writing of the manuscript. The first author focused on the AI mechanisms and mapping them to legal requirements, while the second author focused on researching the judicial side and identifying the most challenging aspects of the legal domain.


\bibliographystyle{IEEEtran}
\bibliography{legalgenai} 

\end{document}